\documentclass{article}
\usepackage{PRIMEarxiv}

\usepackage[utf8]{inputenc} 
\usepackage[T1]{fontenc}    
\usepackage{hyperref}       
\usepackage{url}            
\usepackage{booktabs}       
\usepackage{amsfonts}       
\usepackage{nicefrac}       
\usepackage{microtype}      
\usepackage{xcolor}         
\usepackage{graphicx, subcaption, multirow} 

\title{Measuring Visual Understanding in Telecom domain: Performance Metrics for Image-to-UML conversion using VLMs}

\author{HG Ranjani, 
  Rutuja Prabhudesai*\\
  Ericsson R\&D, Bangalore, India \\
  \texttt{ranjani.h.g@ericsson.com, rutuja.prabhudesai@iiitb.ac.in\thanks{This work was done during the author's internship at Ericsson R\&D.}}  
}

\begin{document}
\maketitle
\begin{abstract}
 Telecom domain 3GPP documents are replete with images containing sequence diagrams.  
 Advances in Vision-Language Large Models (VLMs) have eased conversion of such images to machine-readable PlantUML (\textit{puml}) formats. However, there is a gap in evaluation of such conversions - existing works do not compare \textit{puml} scripts for various components. In this work, we propose performance metrics to measure the effectiveness of such conversions. A dataset of sequence diagrams from 3GPP documents is chosen to be representative of domain-specific actual scenarios. We compare \textit{puml} outputs from two VLMs — Claude Sonnet and GPT-4V — against manually created ground truth representations. We use version control tools to capture differences and introduce standard performance metrics to measure accuracies along various components: participant identification, message flow accuracy, sequence ordering, and grouping construct preservation. We demonstrate effectiveness of proposed metrics in quantifying conversion errors across various components of \textit{puml} scripts. 
 The results show that nodes, edges and messages are accurately captured. However, we observe that VLMs do not necessarily perform well on complex structures such as notes, box, groups. Our experiments and performance metrics indicates a need for better representation of these components in training data for fine-tuned VLMs.
\end{abstract}

\section{Introduction}
Sequence diagrams are widely used to represent signaling sequences and interactions among system components. However, these diagrams are often available only as static images within technical documents and scattered across versions and sections. This limits machine-readability of such sequences and their usability in tools that support analysis, simulation, automated verification and/or troubleshooting. We consider the telecom domain as a case in point to illustrate some challenges using 3rd Generation Partnership Project (3GPP) specifications \cite{3gpp_release_18}. These are publicly available as word documents containing text, tables, equations and images {\color{black}\cite{eval2024rag, table2024eval, speechequations2025}} including sequence diagrams (as images) within to illustrate procedural call flows across various network entities in various scenarios.

Recent advances in Vision-Language Large Models (VLMs) have enabled the extraction of structured information from images, including charts, tables and UML (Unified Modeling Language) diagrams.
Several studies have proposed methods to extract UML components from visual representations, converting image-based diagrams into machine-readable formats using tools such as PlantUML to create \textit{puml} scripts \cite{PlantUML, PlantUML1, romeo2025uml}. These approaches aid towards automating the conversion of legacy diagram archives into usable data. In this work, we use \textit{puml} to refer to UML scripts (accessed or generated using PlantUML tools or equivalent ones). 
\begin{figure*}[t]
    \centering
    \includegraphics[width=0.75\textwidth]{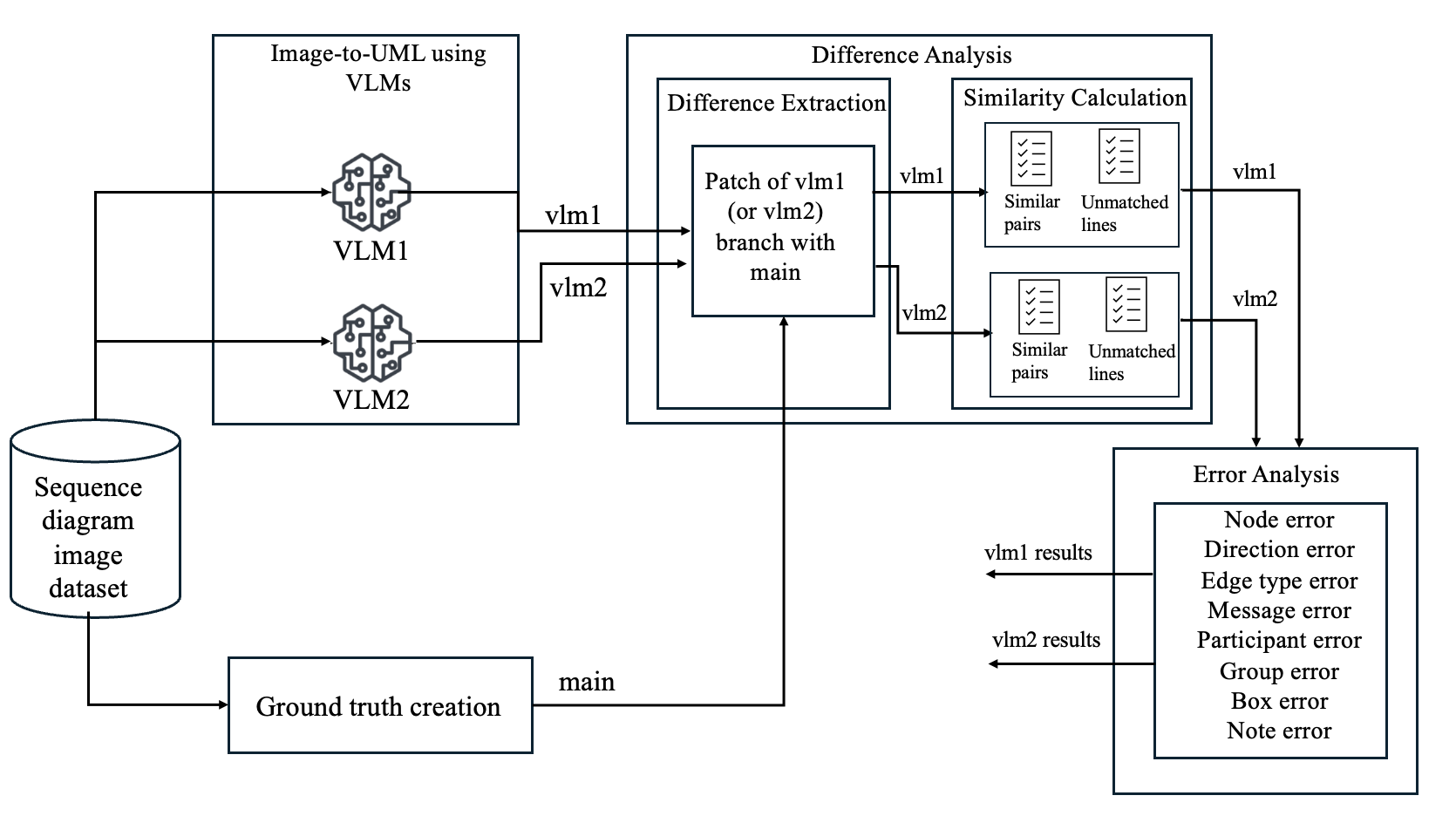}
    \caption{A block diagram depiction of proposed approach to compare image-to-UMLs outputs from two VLMs.}
    \label{fig:proposed_block_diagram}
\end{figure*}
The work in \cite{de2024evaluating} uses LLMs to {\color{black}generate} UML class diagrams. The diagrams are analyzed for syntactic, semantic, and pragmatic quality against that of human generated UML diagrams. In \cite{ye2024beyond}, flowchart images are converted to graphical structures using VLMs. Then, these structures are compared for the optimal representation format (\textit{puml}, Mermaid or Graphviz) for improved performance in reasoning based question answering (QA) task. For evaluating the UML representation format, node-F1 and edge-F1 metrics have been considered. The work of \cite{axt2023transformation} converts human sketches into UML diagrams using OpenCV libraries. The UML diagrams are evaluated {\color{black}based} on precision and recall of classes, inheritances, and associations. The work \cite{conrardy2024image} also addresses converting human sketches into UML diagrams, by using VLMs. The approach is based on chain-of-thoughts (CoT) via multiple prompts. They evaluate the approach through count of number of mistakes (including non-compilation, hallucinations, and similar errors).  Another recent work \cite{bates2025unified} also leverages multi-modal LLMs to convert image based UML diagrams to \textit{puml} format using fine-tuned VLMs. A synthetic image-based UML dataset was created. The generated \textit{puml} was again visualized as an image and compared with the original image for visual fidelity using Structural Similarity Index Measure (SSIM) and use Bilingual Evaluation Understudy (BLEU) scores for semantic similarity.  

Despite these recent efforts, we observe a significant gap in terms of evaluations: currently there is no systematic approach to evaluate the correctness of such image-to-UML datasets. As a result, it is challenging to assess the accuracy of the converted \textit{puml} representation in terms of efficiency of capture of various structural components of the original image. Thus, benchmarking of existing methods and the development of more robust systems is difficult. This is specifically true for complex \textit{puml} representations where synchronous and asynchronous events, grouping of events are important information to be captured.  

A wishlist of error metrics to compare two UML diagrams include errors pertaining to components such as participants, connector types, connector directions, messages passed, notes, sequences, groups amongst other syntactic and semantic components.  In this work, we address this gap by introducing a set of metrics across these components to measure correctness of \textit{puml} conversions. 

The dataset chosen focuses on the telecom domain using 3GPP sequence diagrams (parsed from publicly available documents), where the sequence diagrams range from simple to complex and include synchronous and asynchronous events and contain many \textit{puml} components listed above. We propose to measure the differences between two \textit{puml} representations. Towards this, we manually curate ground truth  and quantify \textit{puml} output VLM  for various components: participants, message flow, ordering, and grouping constructs.

The contributions of this work are: (i) compare VLM (Claude and GPT \footnote{In this work, we use GPT-4 and GPT to refer to GPT4-V. Similarly, we use Claude to refer to Claude Sonnet.}) performances based on their ability to convert sequence diagram images to \textit{puml} format (ii) propose use of version control tools for capture differences between \textit{puml} format (iii) introduce metrics for various components to measure differences between various components of \textit{puml} representations. 

The manuscript is organized as follows: Section \ref{sec:approach} details proposed approach, followed by experimental setup in Section \ref{sec:exptal_setup} which includes dataset preparation, \textit{puml} script generation and difference analysis. This is followed by the detailing of metrics introduced for error analysis in Section \ref{sec:error_analysis}. Detailed analysis of the results is in Section \ref{sec:results} followed by concluding remarks in Section \ref{sec:conclusion}.

\section{Proposed Approach}
\label{sec:approach}

Figure \ref{fig:proposed_block_diagram} shows a block diagram representation of the proposed approach. As can be seen, there are 3 major steps: 
\begin{itemize}
    \item Image-to-UML conversion using VLMs.
   \item Difference analysis: extract the patch files between VLM outputs and ground truth. For analysis, the patch files are grouped into two groups (i) similar pairs and (ii) unmatched lines.
   \item Error analysis: introduce error metrics into various categories such as node-based, edge-based, message-based and other structural components (detailed in Section {\color{black}\ref{ssec:similarity_calculation}})
\end{itemize}

We detail each step in subsections below. 

 \subsection{Image-to-UML conversion using VLMs}
We parse images from the publicly available 3GPP standard documents. We manually categorize these imagesinto two: (i) sequence diagrams and (ii) non-sequence diagrams. This image categorization can be automated with a fine-tuned classifier such as {\color{black}\cite{moreno2020automatic, graphkdd2025}}, but is not the focus of this work. We curate a subset of these sequence diagrams for analysis in this work. More details on the dataset is in Section \ref{ssec:data}.  

We consider two VLMs (Claude and GPT-4) for converting sequence diagram images to \textit{puml} formats. We include details of the VLM models considered in Section \ref{ssec:models}.


\subsection{Difference Analysis}
\label{ssec:diff_analysis_approach}
The \textit{puml} files are diagrams-as-code scripts. In order to capture and evaluate the differences between the VLM outputs and ground truth, we consider version control tools, due to the textual nature of the \textit{puml} scripts. We obtain the differences between the ground truth and VLM outputs by first extracting differences and then performing a similarity calculation. 
\subsubsection{Difference extraction}
There are two components involved in difference extraction:
\begin{itemize}
    \item Repository creation: Here, we consider the ground truth \textit{puml} scripts in the main branch of a git repository while each set of the VLM outputs 
    {\color{black}{with same name as corresponding ground truth files}} 
    can be considered under a separate branch. 
    \item Patch extraction: Git diff or patch files are generated for each VLM model output with respect to main branch. These captures document specific differences with respect to ground truth. 
\end{itemize}

\subsubsection{Similarity calculation}
\label{ssec:similarity_calculation}
We propose a multi-step approach for similarity calculation using the git diff/patch files.
\begin{enumerate}
\item \textbf{Preprocess the diff/patch} file such that only relevant lines are considered for analysis:    \begin{itemize}
        \item Lines containing arrows (connections between participants) such as ``\textit{\textit{PGWA -> SGW : 2a. Update Bearer Request (PGW Change Info)}}''
        \item Lines with keywords such as \texttt{"group"}, \texttt{"note"}, \texttt{"box"}, \texttt{"participant"},\texttt{"actor"}.
        \item Non-structural lines are excluded (e.g., \texttt{"end"}, \texttt{"end group"}, \texttt{"end note"}, \texttt{"end box"})
    \end{itemize}
    \item \textbf{Pairing of lines} to align groups of lines and find optimal matching lines from the diff file. For this, Levenshtein distance {\cite{Levenshtein1966SPhD}} is calculated between every element of removed (starts with -) and added (starts with +) lines to quantify textual differences. Linear sum assignment implementation {\cite{linear_sum_assignment}} of modified Jonker-Volgenant algorithm {\cite{DavidCrouse}} is applied to find optimal matching between removed and added lines, minimizing the total distance. This pairing approach identifies candidate lines in the model output which correspond to similar lines in the ground truth. Similar lines can be considered to be aligned based on minimal Levenshtein distance between candidate pairs. 
    \item \textbf{Post processing }output of the pairing of lines process includes: \begin{itemize}
    \item Unpaired lines from removed groups are considered as elements missing in model output.
    \item Unpaired lines from added groups are considered as insertions w.r.t. ground truth.
   \item Paired lines with differences are categorized either as {\color{black}substitutions or as a combination of additions and deletions, depending on the component, nature and extent of the change.}
\end{itemize}
\end{enumerate}

\subsection{Error Analysis}

Consider a sample \textit{puml} sequence component ``\textit{PGWA -> SGW : 2a. Update Bearer Request (PGW Change Info)}''. Here, `PGWA' and `SGW' are considered as nodes, `->' corresponds to edges, `2a. Update Bearer Request (PGW Change Info)' corresponds to message element. In addition, there are other components such as box, group, participant and notes. 

In this subsection, we describe the process considered for error categorization into various components, followed by the metrics introduced in this work to measure the categorized errors.

\subsubsection{Error Categorization}
\label{ssec:error_categorization}

For each paired line with Levenshtein distance $> 0$ (indicating a difference), and unpaired lines, the specific type of difference is determined through regex pattern matching and context analysis (such as presence of ``->", ``:" patterns). Each difference is categorized into one of the pre-defined categories based on the specific nature of the error (e.g., arrow direction, message content, participant name).
The categories are based on the specific element types:
    \begin{itemize}
        \item Node related errors (participant identification issues)
        \item Edge or connection errors (arrow direction, type)
        \item Message content errors
        \item Other structural element errors: notes, groups, boxes, participants\footnote{We differentiate nodes and participants based on the context of their occurrence. Lines such as `participant PGWA' contribute to participant category while lines such as `PGWA -> SGW : 2a. Update Bearer Request' contributes to node category.}.
    \end{itemize}

\subsubsection{Error Metrics}
\label{sec:error_analysis}
We quantify the differences between the ground truth and model output scripts for each of the categorized errors using the metrics introduced and detailed component-wise below:  
\begin{itemize}
    \item \textbf{Node related Metrics}: These metrics are closely associated with participants occurring in each sequence of the \textit{puml} scripts. 
\begin{itemize}
     \item \textit{Node Insertion rate}: Count of nodes in present in model output, but not in ground truth divided by total number of nodes in ground truth. 
    \item \textit{Node Deletion rate}: Count of nodes not present in model output, but present in ground truth divided by total number of nodes in ground truth.
    \item \textit{Node Substitution rate}: Count of nodes with incorrect naming/representation divided by total number of nodes in ground truth. {\color{black}It is also associated with edit distance to quantify the incorrectness}. 
\end{itemize}
\item \textbf{Edge/Connection Metrics}: These metrics are associated with connectors (or edges)  and include: 
\begin{itemize}
    \item \textit{Edge Direction change rate}: Count of arrows with incorrect direction in model output divided by total number of arrows in ground truth. 
    \item \textit{Edge Direction insertion rate}: Count of inserted arrows not in ground truth, but present in model output divided by total number of arrows in ground truth. 
    \item \textit{Edge Direction deletion rate}: Count of deleted arrows not in model output, but present in ground truth divided by total number of arrows in ground truth.
    \item \textit{Edge Direction type change rate}: Count of arrows with incorrect type divided by total number of arrows in ground truth (e.g., solid vs. dashed i.e., `->' vs. `- - >' representing synchronous message vs. asynchronous message)
    \end{itemize}
 \item \textbf{Message related Metrics}: Most sequence diagrams considered show passing of messages between participants. Through these metrics, we can measure correctness of messages passed between participants.
    \begin{itemize}
   \item \textit{Message insertion rate}: Count of inserted messages present in model output, not present in ground truth divided by total number of messages in ground truth. 
    \item \textit{Message deletion rate}: Count of messages present in ground truth, not present in model output divided by total number of messages in ground truth. 
    \item \textit{Message change rate}: Count of messages with non-exact matches in model output divided by total number of messages in ground truth. 
\end{itemize}
\item \textbf{Structural Element Metrics}: In addition to the nodes (participants), edges (connectors) and messages, there exist other structural elements in a complex \textit{puml} diagram such as notes, groups, boxes. 
\begin{itemize}
   \item \textit{Note Changes}: Rate of insertion, deletion, and substitutions of notes.
    \item \textit{Group Changes}: Rate of insertion, deletion, and substitutions of groups.
    \item \textit{Box Changes}: Rate of insertion, deletion, and substitution of boxes.
    \item \textit{Participant Changes}: Rate of insertion, deletion, and substitution of participants.
\end{itemize}
\end{itemize}

\section{Experimental Setup}
\label{sec:exptal_setup}
In this section, we detail the setup considered for the experiments to measure effectiveness of VLMs for sequence diagram images-to-UML conversion. 
\subsection{Dataset Preparation}
\label{ssec:data}
 We parse 3GPP (Rel 18) documents \cite{3gpp_release_18} for all the images in the word doc and docx files. The corresponding image-dataset comprises of $\sim$14000 images. The images along with their captions are collected and labeled in accordance to the order of their occurrence in the documents. This dataset contains various categories corresponding to graphs, sequence diagrams, frequency diagrams, block diagrams and schematic diagrams. {\color{black}A sample set is shown in Figure \ref{fig:3gpp_img_categories}} in Appendix \ref{apdx:3gpp_image_categories}.

These images are manually classified into sequence and non-sequence diagram categories; $32$\% of the images are  sequence diagrams. This sequence diagram dataset, along with its corresponding captions corresponds to {\color{black}4010} images. The total pixel count of these images ranges between {\color{black} \(240 \times 57\)} to  {\color{black}\( 7548 \times 6510\)}.
This collection of sequence diagrams forms the dataset considered for further steps.
We highlight that these images do not have the ground truth \textit{puml} script readily available. 

A sample representative subset of 50 sequence diagrams are selected from the complete dataset to create ground truth \textit{puml} files. 
{\color{black}The selection criteria includes diversity of diagram features, including arrow types and styles, color schemes, note positioning, special features such as loops, alternative paths and participant representation styles}. All results in this work pertain to these 50 sequence diagrams. 

Although readers might presume that 50 files is a modest size dataset, we would like to highlight that the purpose of this work is to propose evaluation metrics considering associated complexities in comparing two \textit{puml} scripts than evaluate the VLMs themselves on large datasets. 

\subsection{PlantUML Script Generation}
Here, we describe the ground truth preparation and approach for \textit{puml} script generation using VLMs. 
\subsubsection{Ground Truth}
The ground truth \textit{puml} scripts are manually created for all 50 selected images.
The resulting ground truth scripts serve as the reference for evaluation.  The overall number of lines in ground truth \textit{puml} script corresponds to {$\sim 2500$}. The distribution of 50 files w.r.t. number of lines in ground truth is shown in Table \ref{tab:stats_lines}.
\begin{table}[h]
    \centering
    \begin{tabular}{|c|c|}
    \hline
       Range of \textit{puml} script lines  & Count of \textit{puml} files \\\hline
       1-20  & 10 \\
       21-30 & 13\\
       31-40 & 5\\
       41-50 & 13 \\
       51-100 & 9 \\
         \hline 
    \end{tabular}
    \caption{Distribution of \textit{puml} files w.r.t. number of lines of script in ground truth.}
    \label{tab:stats_lines}
\end{table}

\subsubsection{VLM Prompt}
The following prompt was used to generate puml scripts from the diagram images using VLM:
\begin{quote}
\textit{
Generate puml script for given 3GPP standard call flow diagram of "\{Caption of the image\}" according to puml documentation. Please consider following important points:\\
1. Correctly identify participants/actors.\\
2. Correctly identify the connection between the nodes using given arrows.\\
3. Correctly identify the arrow direction, start and end of the arrow.\\
4. Correctly identify text associated to the arrow.\\
5. If any text is in rectangles consider them as notes and write them in appropriate place.\\
6. Give numbering to each call sequentially.\\
7. Correctly identify color if any.}
\end{quote}
\subsubsection{VLM Models}
\label{ssec:models}
The prompt described above was used with two VLMs: 
    \begin{itemize}
        \item Claude 3.7 Sonnet model from Anthropic {\color{black}\cite{Claude}}
        \item GPT-4-Vision model from OpenAI {\color{black}\cite{GPT4}}
    \end{itemize}
VLMs is an evolving field with new models released quite frequently. At this juncture, we again highlight that although other VLMs can be considered for comparison, the focus of this work is to establish an approach to evaluate \textit{puml} scripts generated from VLMs and not to evaluate all the VLMs as such. 

\begin{figure}[h]
    \centering
    \includegraphics[width=\linewidth]{ 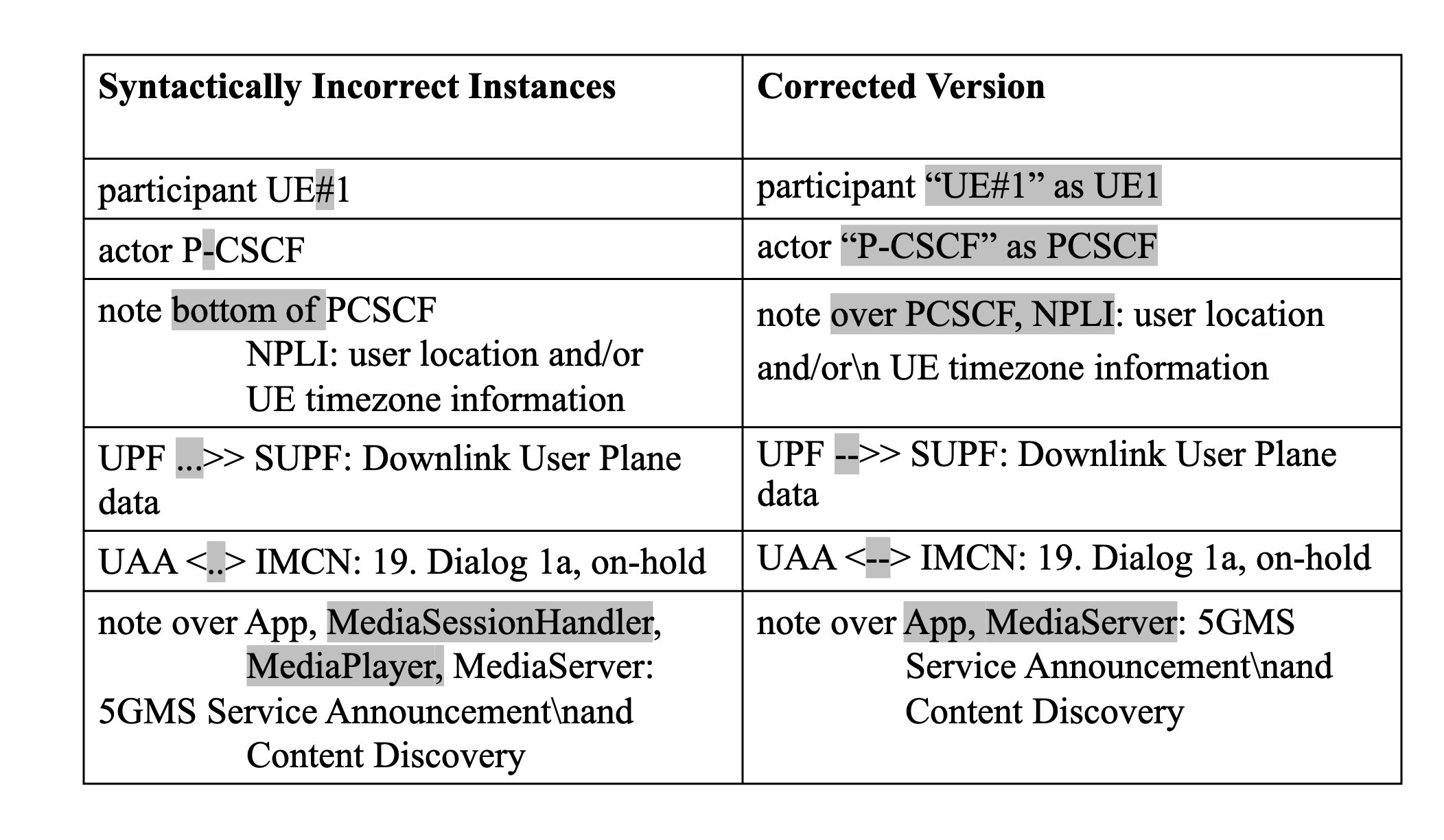} 
    \caption{Sample snapshot related to manual syntax corrections on VLM outputs with the errors and corrections highlighted in gray.}
    \label{fig:syntax_correction}
\end{figure}

The generated \textit{puml} scripts are rendered using the \textit{puml} web server to manually visually verify syntactic correctness. We do not penalize VLM model outputs unnecessarily during error analysis. Hence, in scenarios where the scripts were not syntactically correct (and leads to not being able to generate the \textit{puml} image), we identify and rectify minor issues such as introduction of spurious characters such as `\#' and `-', replace elements identified as actors (by VLMs) as participants because the ground truth contains only participants, invalid arrow syntax such as `..>' to $-->$. In addition, unsupported note placements, overuse of participants in note overs are manually corrected and not counted towards errors. A few sample instances of such corrections (not counted towards errors) are depicted in {\color{black} Figure \ref{fig:syntax_correction}}. 

Figure~\ref{fig:fullpage_grid} in Appendix~\ref{apndx:sequence_diagram_images} depicts a sample sequence diagram image from the 3GPP standard, along with its corresponding visual renderings generated from \textit{.puml} scripts (shown in Figure~\ref{fig:pumlscripts}) including the ground truth, Claude output, and GPT-4 output.
{\color{black}{We have included manually inserted annotations to indicate definitions of some of the proposed metrics from the \textit{puml} scripts.}}

\subsection{Difference Analysis}
\label{ssec:diff_analysis_setup}
We detail the experimental setup for repository creation and for patch extraction steps corresponding to Section \ref{ssec:diff_analysis_approach} here. 

\subsubsection{Repository Structure}
A Git repository is created to manage the different versions of \textit{puml} scripts. Each sequence diagram image is converted to it's corresponding \textit{puml} file within a folder corresponding to the document name. 
\begin{figure}[t]
    \centering
    \includegraphics[width=\linewidth]{ 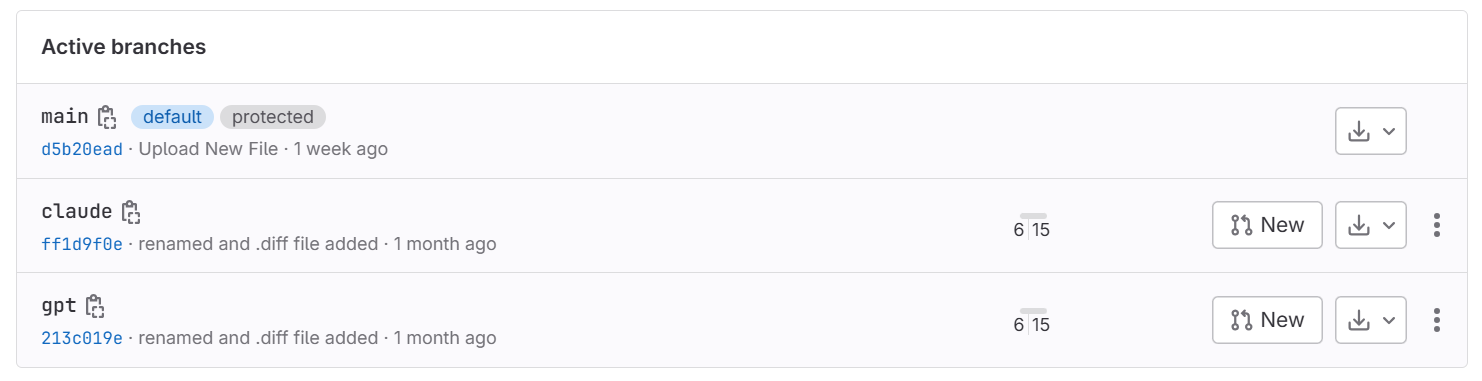} 
    \caption{Snapshot of repository with three branches - main, claude and gpt}
    \label{fig:repo_Structure}
\end{figure}
Three branches are created:
\begin{itemize}
       \item \texttt{main}: containing the manually verified ground truth \textit{puml} scripts
       \item \texttt{claude}: containing Claude generated \textit{puml} scripts
       \item \texttt{gpt}: containing GPT-4 generated \textit{puml} scripts
\end{itemize}
Figure  \ref{fig:repo_Structure} shows a sample snapshot of repository branches.
\subsubsection{Difference Extraction}
\begin{figure*}[!htbp]
    \centering
    \begin{minipage}[b]{0.48\textwidth}
        \centering
        \includegraphics[width=\linewidth]{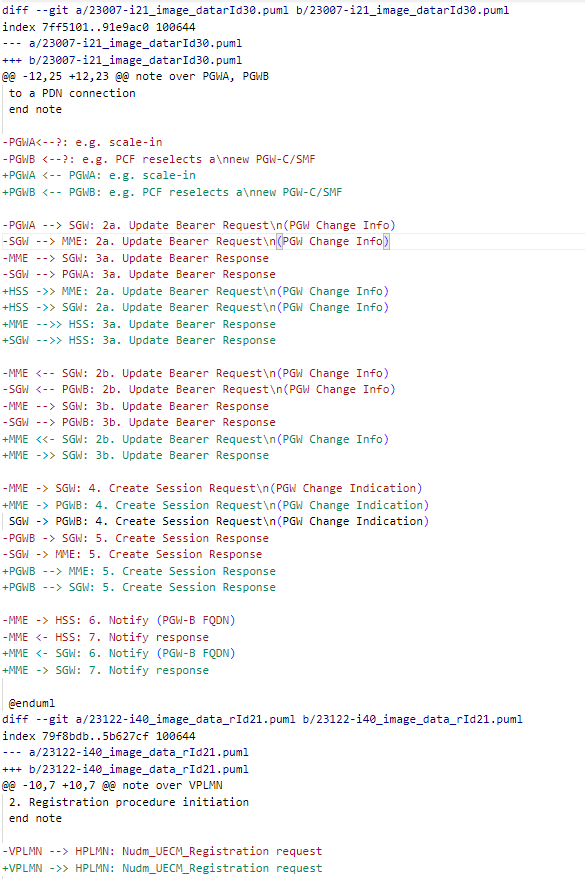}
        \caption{An example of patch/diff file obtained by comparing model (claude/gpt branch) output with ground truth (main branch).}
        \label{fig:patch}
    \end{minipage}
    \hfill
    \begin{minipage}[b]{0.48\textwidth}
        \centering
        \includegraphics[width=0.72\linewidth]{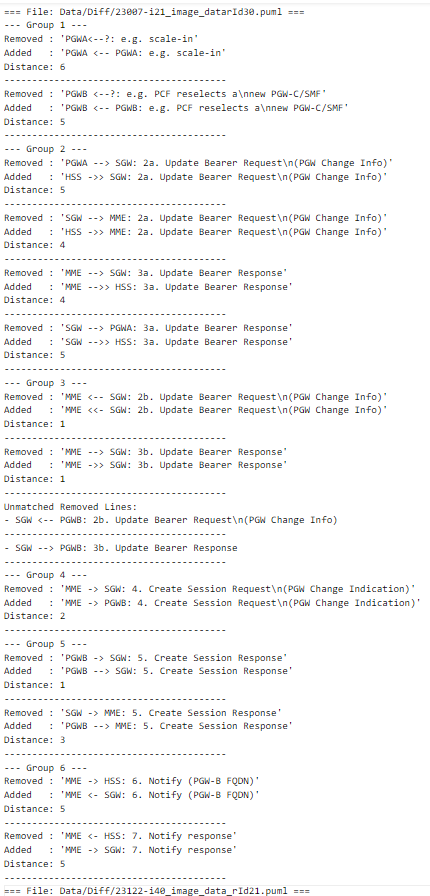}
        \caption{An example of distance calculation and classification of matched pairs and unmatched lines from patch/diff file shown in Fig.~\ref{fig:patch}.}
        \label{fig:Difference}
    \end{minipage}
\end{figure*}

A Git diff analysis is performed to identify differences between models and ground truth. Towards this, two sets of patch/diff files are generated for each comparison to document specific differences:
\begin{itemize}
    \item \texttt{main} (ground truth) with \texttt{claude} branch
    \item \texttt{main} (ground truth) with \texttt{gpt} branch
\end{itemize}
These patch/diff files are used for capturing and quantifying differences between ground truth and model outputs. A snapshot of diff file is shown in Figure \ref{fig:patch}. Figure \ref{fig:Difference} depicts an example of matched and unmatched lines from the depicted patch file .

\begin{table*}[!hbp]
\centering
\renewcommand{\arraystretch}{1.3}
\resizebox{\textwidth}{!}{
\begin{tabular}{|l|r|r|r|r|r|r|r|r|}
\hline
\textbf{} & \textbf{Node} & \textbf{Direction change} & \textbf{Direction type} & \textbf{Message} & \textbf{Box} & \textbf{Group} & \textbf{Note} & \textbf{Participants} \\
\hline\hline
\multicolumn{9}{|c|}{Ground truth count}\\
\hline
 & 1736 & 881 & 881 & 873 & 19 & 39 & 229 & 278 \\\hline\hline
\multicolumn{9}{|c|}{Error analysis for Claude output}\\
\hline
\textbf{Insertion (\%)} & 13.02 & 12.71 & 0.00 & 13.06 & 42.11 & 0.00 & 17.03 & 0.72 \\
\hline
\textbf{Deletion (\%)} & 15.78 & 15.44 & 0.00 & 14.89 & 52.63 & 69.23 & 17.90 & 2.88 \\
\hline
\textbf{Substitution (\%)} & 12.56 & 6.02 & 10.90 & 11.23 & 21.05 & 15.38 & 31.88 & 8.63 \\
\hline
\hline
\multicolumn{9}{|c|}{Error analysis for GPT-4 output}\\\hline
\textbf{Insertion (\%)} & 19.76 & 19.41 & 0.00 & 19.70 & 0.00 & 69.23 & 42.36 & 1.44 \\
\hline
\textbf{Deletion (\%)} & 18.66 & 18.50 & 0.00 & 17.87 & 100.00 & 76.92 & 64.19 & 5.76 \\
\hline
\textbf{Substitution (\%)} & 34.10 & 7.95 & 16.00 & 39.29 & 0.00 & 5.13 & 35.37 & 40.64 \\
\hline
\end{tabular}
}
\caption{Statistics of components of \textit{puml} in Ground truth and error analysis metrics for the same using Claude and GPT-4 output files.}
\label{tab:claude_gpt4_results}
\end{table*}
\begin{figure*}[h!]
    \centering
    \begin{minipage}[t]{0.48\textwidth}
        \centering
        \includegraphics[width=\linewidth]{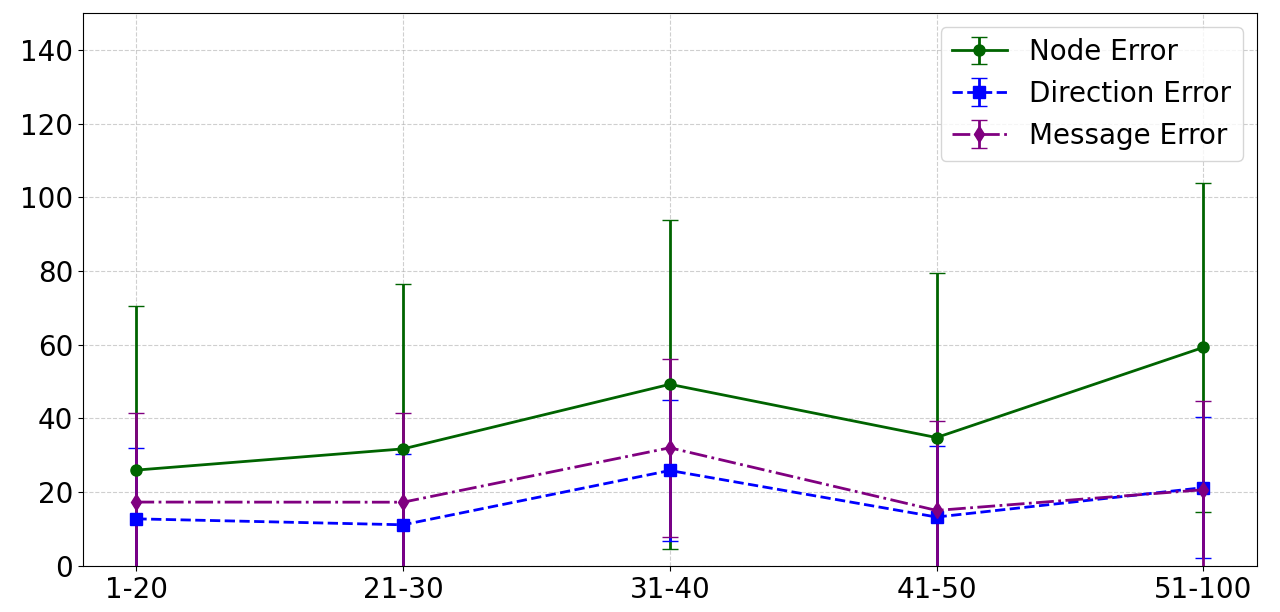}
        \caption{Error bars for node, message, direction error rates in Claude outputs based on number of lines in ground truth files.}
        \label{fig:error_claude}
    \end{minipage}
    \hfill
    \begin{minipage}[t]{0.48\textwidth}
        \centering
        \includegraphics[width=\linewidth]{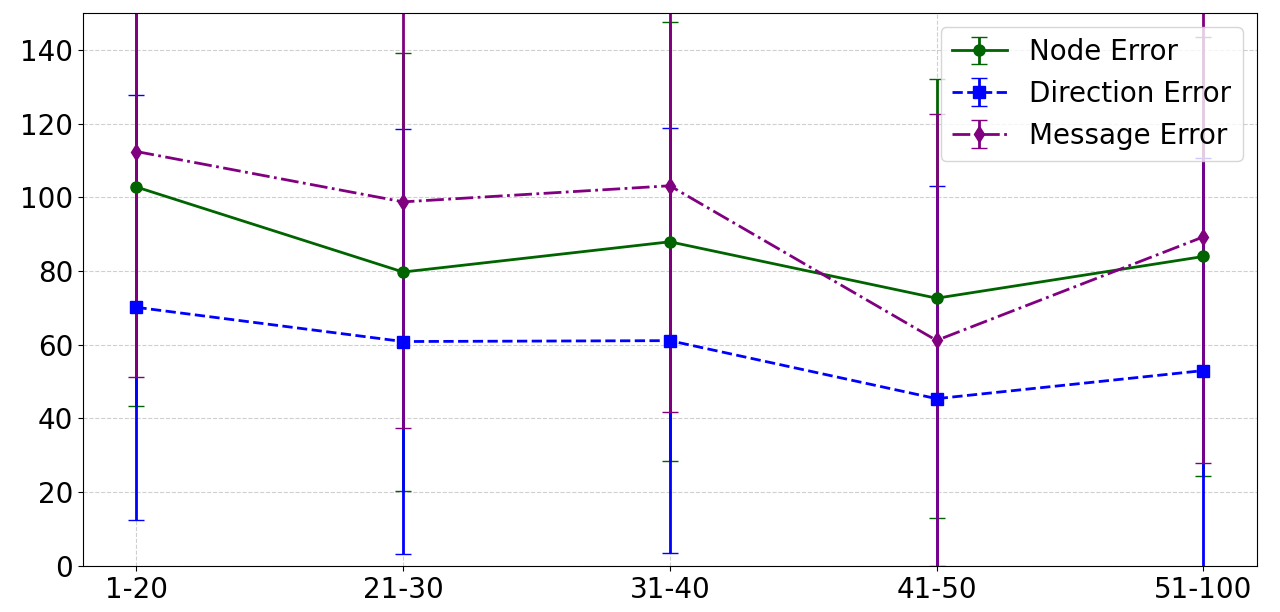}
        \caption{Error bars for node, message, direction error rates in GPT-4 outputs based on number of lines in ground truth files.}
        \label{fig:error_gpt}
    \end{minipage}
\end{figure*}


\section{Results and Analysis}
\label{sec:results}
 
We report the metrics considered for comparing \textit{puml} files, aggregate them and analyze for quantifying behavior of models. 

For the error analysis, we first aggregate error count at both file level and at overall dataset level. 

The overall dataset analysis provides a complete view of the model performance, while the file level analysis can provide more details of when the model doesn't perform well. For each file comparison, the following were calculated:

\begin{itemize}
   \item Total count of elements in the ground truth (nodes, arrows, messages, notes, etc.)
   \item Raw counts of each error type (additions, deletions, substitutions) by category
    \item Percentage of each error type relative to the total count of relevant elements
    \item Error density per diagram (errors per element)
\end{itemize}

Table \ref{tab:claude_gpt4_results} depicts the percentage of insertion, deletion and substitution rates measured for Claude and GPT-4 models w.r.t. the ground truth. We observe that Claude outputs have lesser number of insertion, deletion and substitution rate than GPT-4 outputs across almost all components of \textit{puml}.
The direction type errors are mostly related to substitution because they correspond to synchronous being categorized as asynchronous or vice-versa. 

It is worth noting that there are higher errors in both VLMs outputs with respect to structural elements such as box, group and notes. This indicates that it might be required to fine-tune VLMs for such tasks to reduce error rate across these components. 

We further analyze the VLM outputs using these metrics in terms of number of lines of the \textit{puml} script.  The errors are accumulated across insertion, deletion and substitution categories at a file level and calculated as percentage of total. Figure \ref{fig:error_claude} and Figure \ref{fig:error_gpt} depict the same. {\color{black} We observe that the trend of percent of error in Claude increases with increasing number of lines of script. This is expected because when the sequence diagram is longer, there is less likelihood of retaining the visual context and it is possible that there are more errors. 
With GPT-4, however, the error rate shows a decreasing trend. This, although is not intuitive, hints that GPT-4 retains higher visual context in more complex sequence diagrams over that of simpler one. This needs further investigation. 
In summary, the overall performance seen from Claude model is much better than that of GPT4. }

\section{Conclusions}
\label{sec:conclusion}
It is possible to convert images to \textit{puml} scripts. We have explored the use of VLMs on limited set of sequence diagrams from publicly available 3GPP documents. These have applications in telecom network analysis, simulation, and automated verification systems. 

In this work, we highlight the lack of systematic evaluation of image-to-UML conversion using VLMs. We propose to use version control tools to capture the differences in \textit{puml} representations between ground truth and VLM outputs. We analyze the patch files, align them to be able to capture effectiveness of the \textit{puml} conversion. We propose a set of performance metrics to measure the effectiveness of image-to-uml conversion across various components (\textit{viz.} nodes, edges, messages, participants, box, groups and notes). 
We observe that Claude model is more effective than GPT-4 in the \textit{puml} conversion for the considered dataset. 

The errors are concentrated on complex components such as box, groups, notes. It is expected that a fine-tuning of VLMs focused on sequence diagrams to improve effectiveness for such components. To realize the same, it is important to ensure that training set has these components included appropriately. 

We also observe that errors for Claude increases with increasing number of lines in the script. This is expected as retaining longer visual context may be challenging. However, GPT-4 shows that performance is not much impacted by the number of lines in the script. This is unexpected and necessitates a detailed further analysis. 

In this work, we have focused on simple prompts for the VLMs. Future experiments can include advanced prompts to introduce chain-of-thought approach for image-to-UML conversions. While this work has used limited and focused number sequence diagrams from publicly available 3GPP specifications, the proposed set of performance metrics are agnostic to the domain, source and dataset size of sequence diagram images. 
\bibliographystyle{unsrt}
\bibliography{bib-arxiv}

\appendix

\section{3GPP image categories}
\label{apdx:3gpp_image_categories}
\begin{figure*}[hbt!]
    \centering
    \includegraphics[width=0.75\linewidth]{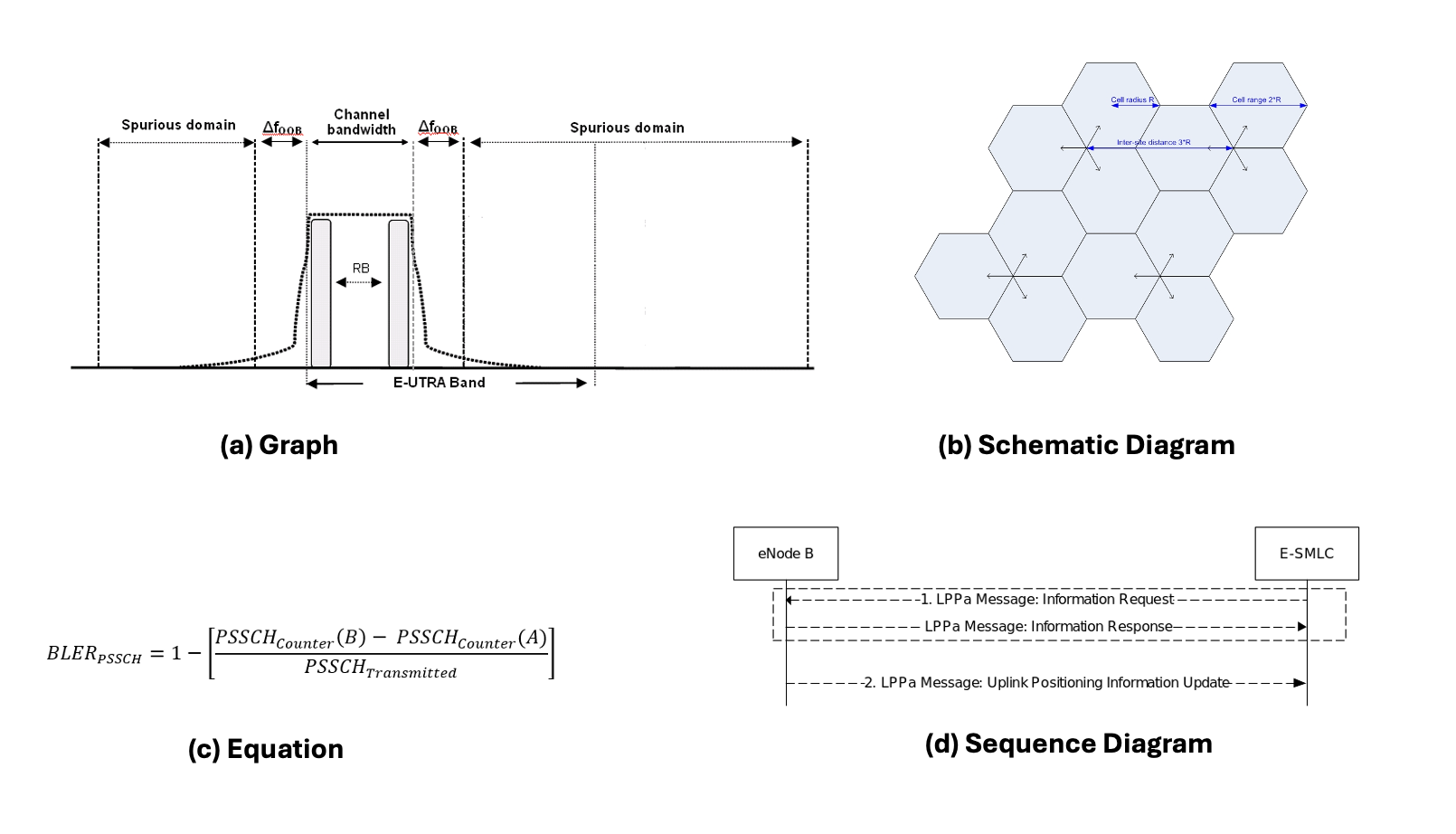}
\caption{Representative images from various categories from the 3GPP dataset.}
\label{fig:3gpp_img_categories}
\end{figure*}

A sample snapshot of various categories of images present in 3GPP standards are shown in Figure \ref{fig:3gpp_img_categories}. 

\section{Sample sequence diagrams from 3GPP specifications}
\label{apndx:sequence_diagram_images}
\begin{figure*}[hb]
\centering
\begin{tabular}{|c|c|c|}
\hline
\includegraphics[height=7cm]{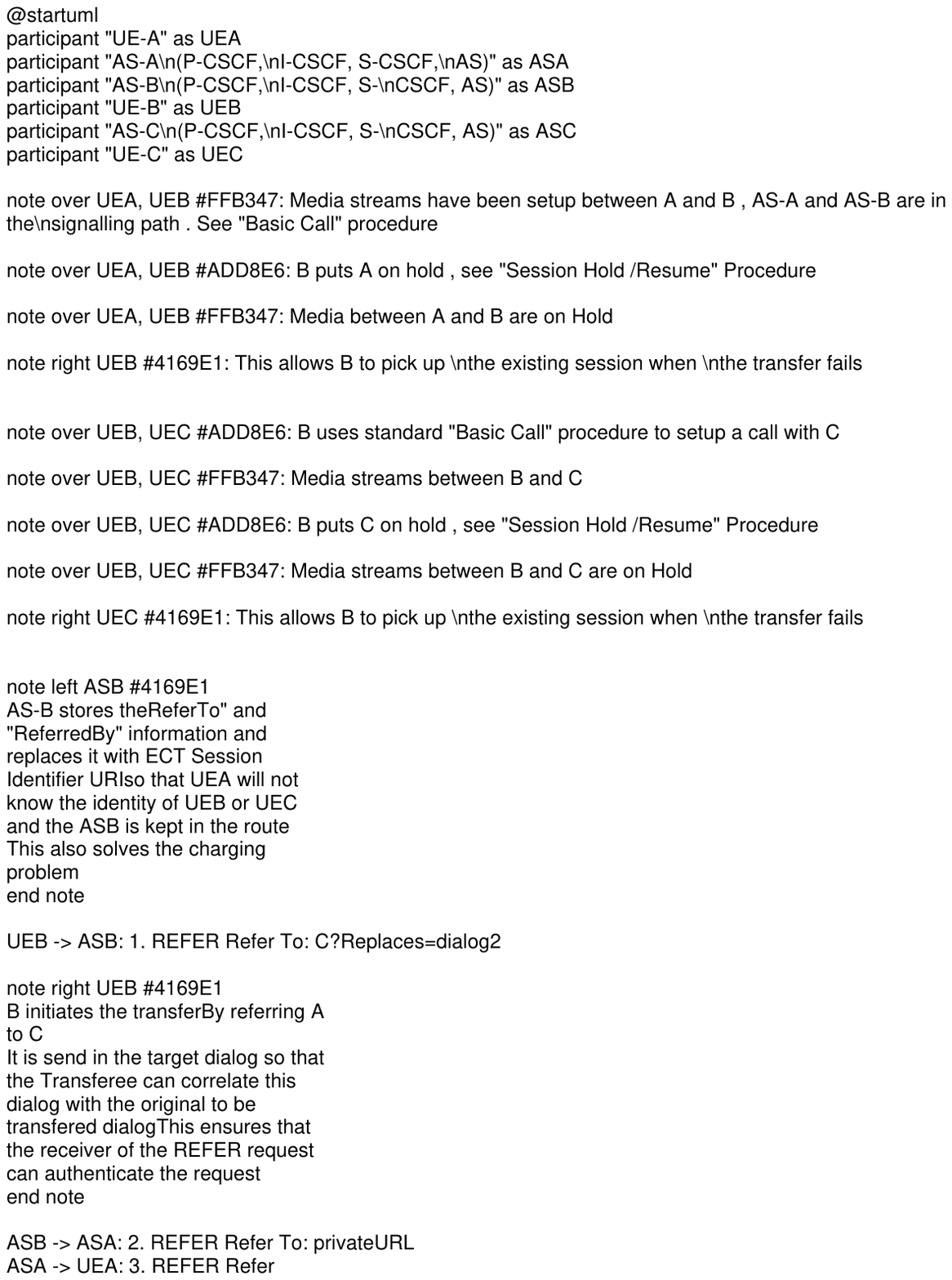} &
\includegraphics[height=7cm]{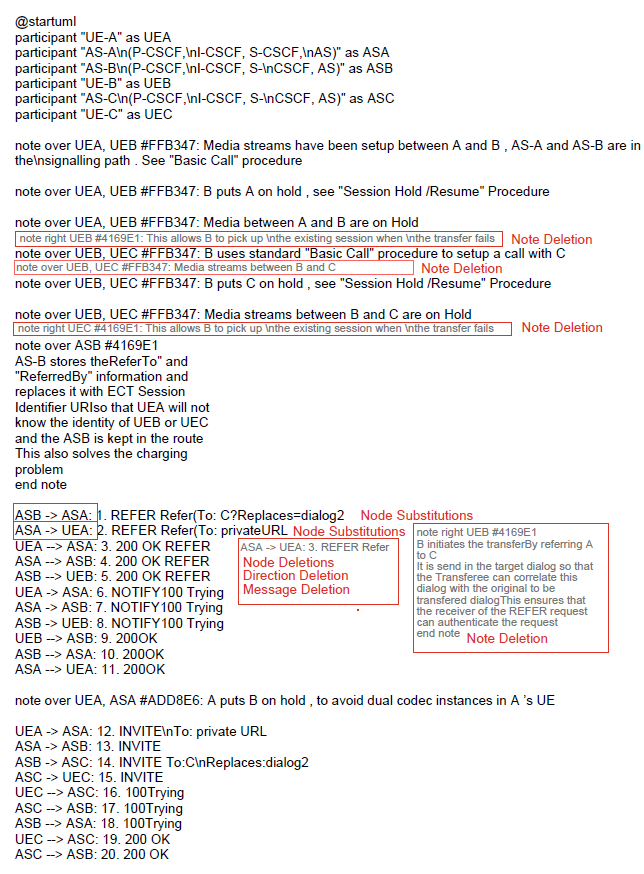} &
\includegraphics[height=7cm]{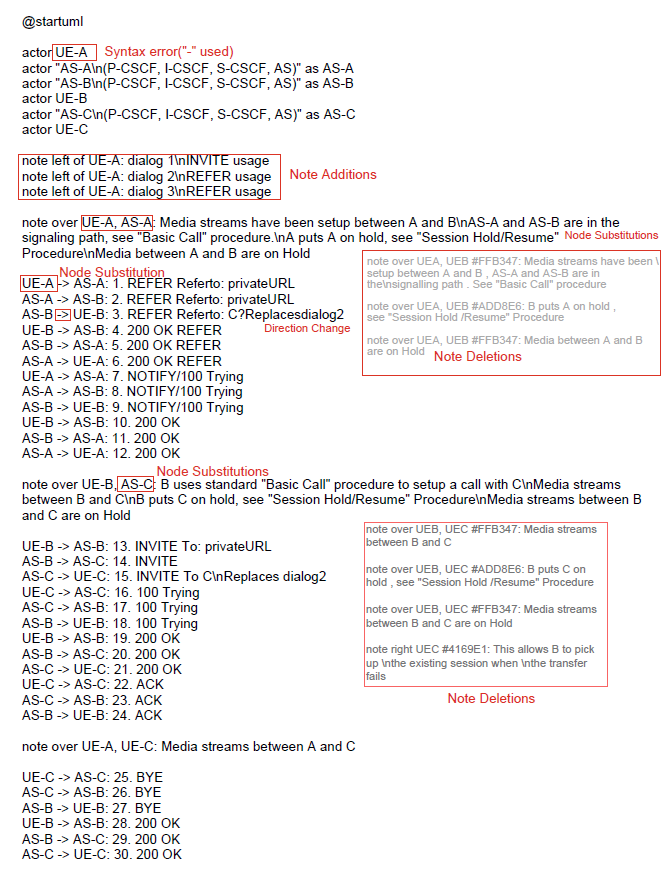} \\
\hline
\end{tabular}
\caption{Comparison of ground truth, Claude and GPT-4 \textit{puml} script with metrics}
\label{fig:pumlscripts}
\end{figure*}

\begin{figure*}[!hbp] 
  \centering
  \begin{minipage}{0.44\textwidth}
    \centering
    \includegraphics[height=0.44\textheight, keepaspectratio]{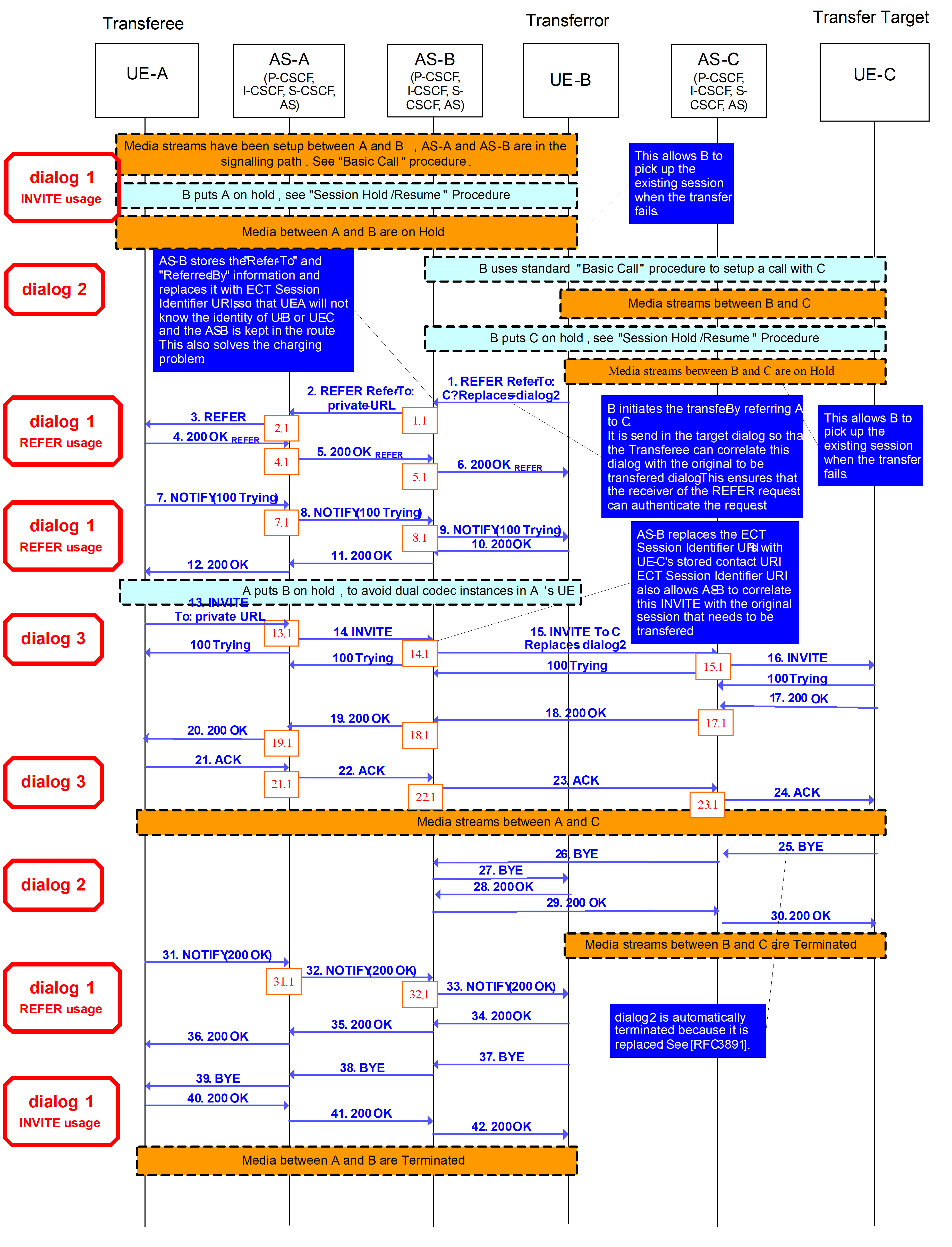}
    \subcaption{Reference image}
  \end{minipage}
  \hfill
  \begin{minipage}{0.44\textwidth}
    \centering
    \includegraphics[height=0.44\textheight, keepaspectratio]{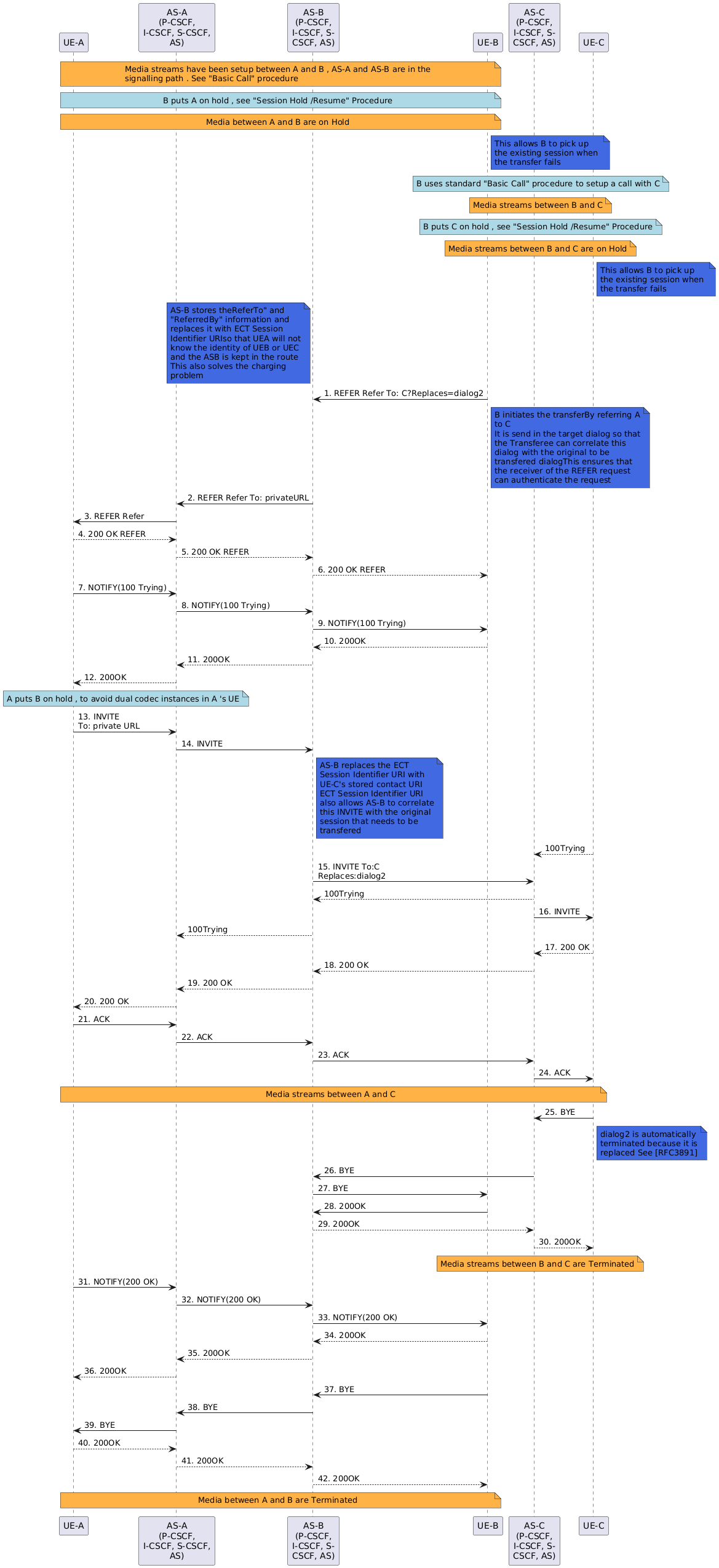}
    \subcaption{Ground Truth}
  \end{minipage}
  
  \vspace{1em}

  \begin{minipage}{0.45\textwidth}
    \centering
    \includegraphics[height=0.45\textheight, keepaspectratio]{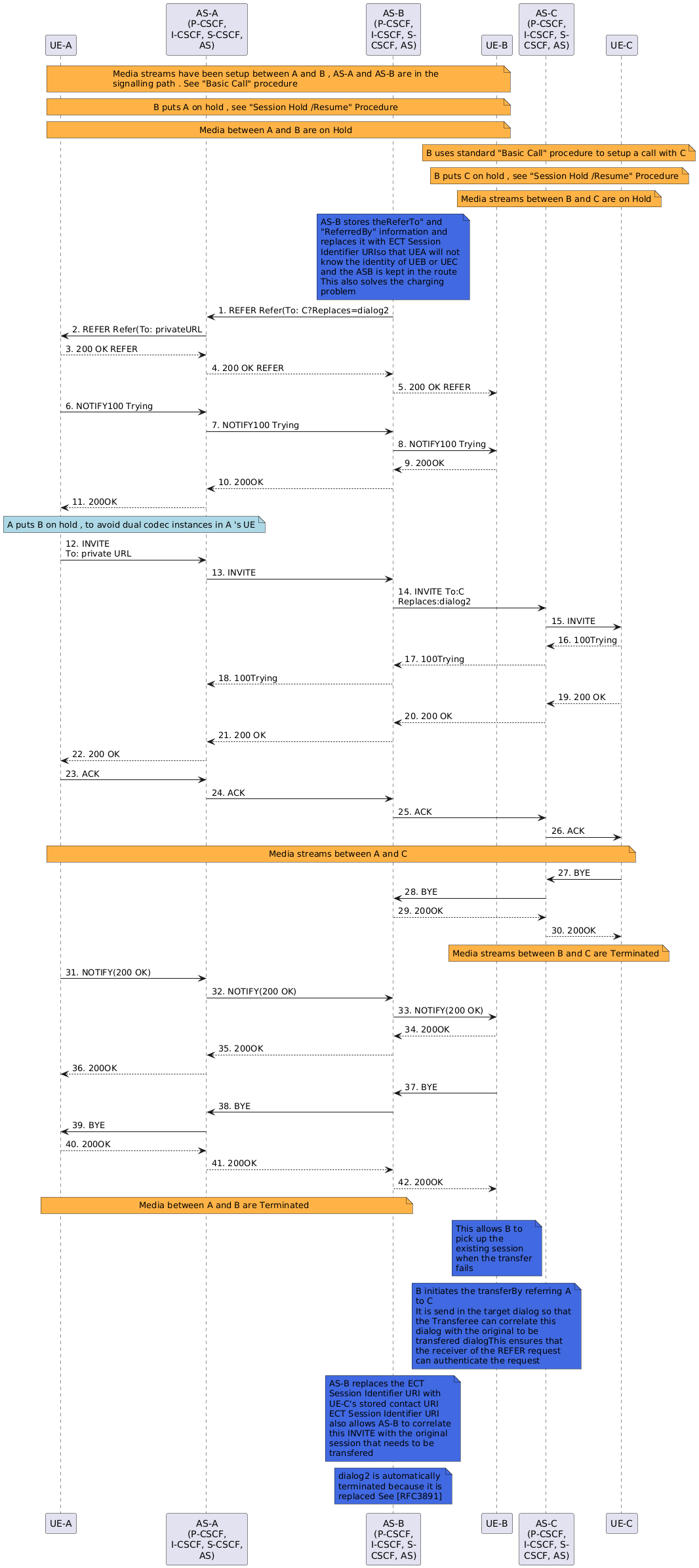}
    \subcaption{Claude Output}
  \end{minipage}
  \hfill
  \begin{minipage}{0.45\textwidth}
    \centering
    \includegraphics[height=0.45\textheight, keepaspectratio]{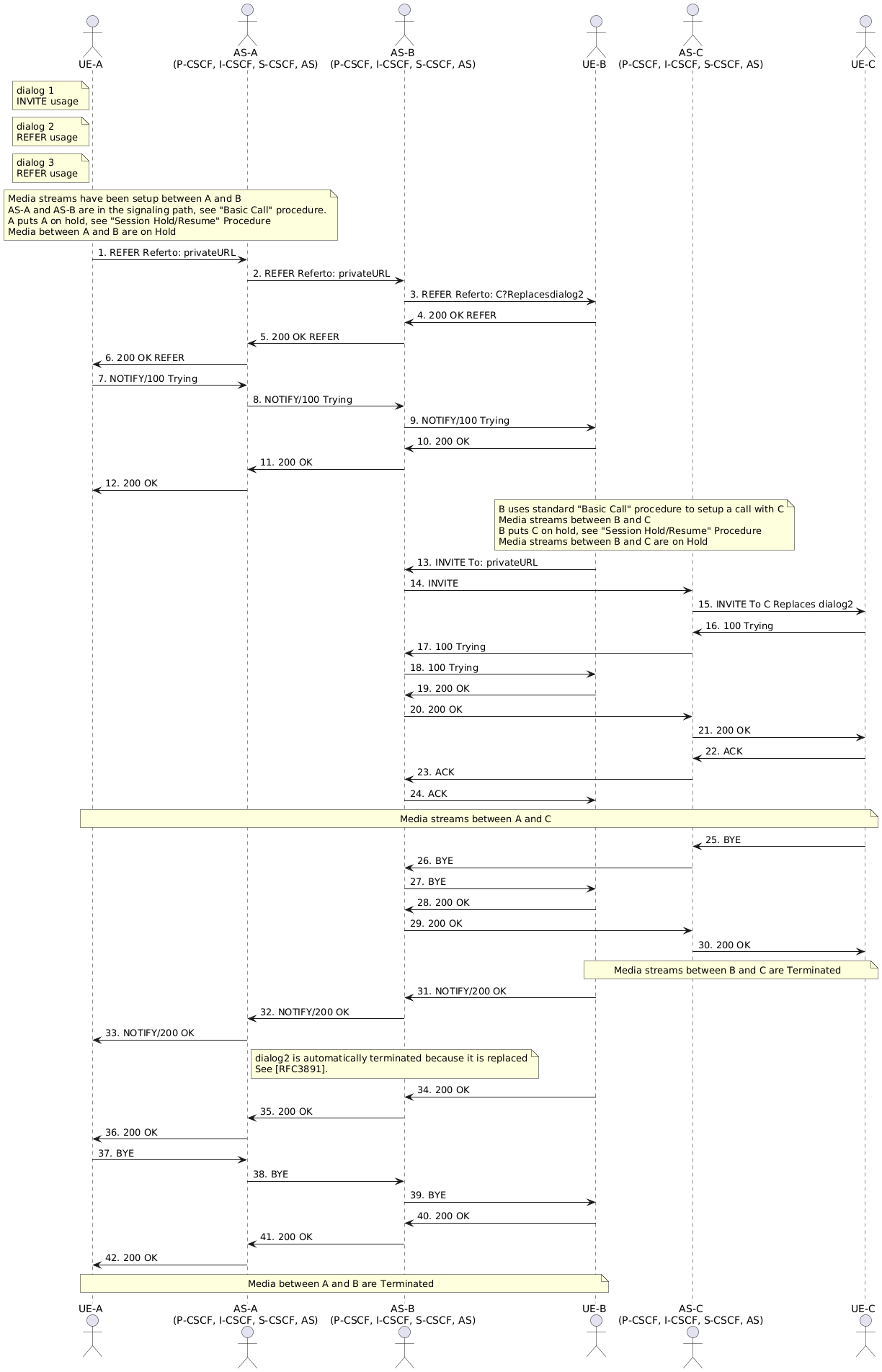}
    \subcaption{GPT-4 Output}
  \end{minipage}

  \caption{Reference image from the 3GPP standard dataset, along with manually created ground truth and outputs from Claude and GPT-4, as visualized on the puml web server.}
  \label{fig:fullpage_grid}
\end{figure*}

Figure \ref{fig:fullpage_grid} shows sample sequence diagram seen in 3GPP standard and it's equivalent image constructed from various puml scripts (viz. ground truth scripts and scripts from two VLMs models as output.). Figure \ref{fig:pumlscripts} shows the corresponding \textit{puml} scripts.

\end{document}